\documentclass{article}

\usepackage[final]{neurips_2019}

\usepackage[english]{babel}
\usepackage[utf8x]{inputenc}
\usepackage{amsmath}
\usepackage{graphicx}
\usepackage[colorinlistoftodos]{todonotes}
\usepackage{caption}

\usepackage[export]{adjustbox}
\usepackage{geometry}
\usepackage{amssymb}
\usepackage{amsmath}
\usepackage{adjustbox}
\usepackage{soul}
\usepackage{multicol}
\usepackage{calc}
\usepackage{floatrow}
\usepackage{listings}
\usepackage{gensymb}  
\usepackage{rotating}
\usepackage{hhline}
\usepackage{multirow}
\usepackage{wrapfig}
\usepackage{booktabs}

\setcitestyle{numbers,square}
\title{Prediction of GNSS Phase Scintillations: A Machine Learning Approach}

\author{%
  Kara Lamb\thanks{Equal Contributions} \\
  Cooperative Institute for Research in the Environmental Sciences/NOAA\\
  Boulder, CO, USA\\
  \And 
  Garima Malhotra*  \\
  University of Michigan\\
  Ann Arbor, Mi, USA\\
  \And
  Athanasios Vlontzos* \\
  Imperial College London \\
  London, UK\\
  \And
  Edward Wagstaff*\\
  University of Oxford \\
  Oxford, UK\\
  \And 
  Atılım Günes Baydin \\ University of Oxford\\ Oxford, UK
  \And 
  Anahita Bhiwandiwalla \\ Intel AI Lab \\ Santa Clara , CA, USA
  \And 
  Yarin Gal \\ University of Oxford \\ Oxford, UK
  \And
  Alfredo Kalaitzis \\ Element AI \\ London, UK
  \And
  Anthony Reina \\ Intel AIPG \\ San Diego, CA, USA\\
  \And
  Asti Bhatt \\ SRI International \\ Menlo Park, CA, USA\\
}

\begin{document}

\maketitle

\begin{abstract}
  A Global Navigation Satellite System (GNSS) uses a constellation of satellites around the earth for accurate navigation, timing, and positioning. Natural phenomena like space weather introduce irregularities in the Earth's ionosphere, disrupting the propagation of the radio signals that GNSS relies upon. Such disruptions affect both the amplitude and the phase of the propagated waves. No physics-based model currently exists to predict the time and location of these disruptions with sufficient accuracy and at relevant scales. In this paper, we focus on predicting the phase fluctuations of GNSS radio waves, known as \emph{phase scintillations}. We propose a novel architecture and loss function to predict $1$ hour in advance the magnitude of phase scintillations within a time window of $\pm 5$ minutes with state-of-the-art performance. 
\end{abstract}

\section{Introduction}
A global navigation satellite system (GNSS) is a constellation of satellites around the Earth used for accurate navigation, timing and positioning. GNSS refers to a large category of commercial products like the Global Positioning System (GPS). As a society, we are becoming increasingly dependent on GNSS technology, with a recent study estimating losses of \$1 billion per day in case of an extended outage~\cite{NISTR2019}. Therefore it becomes imperative to predict disruptions to GNSS signals with good spatial and temporal resolution.

GNSS signals are high frequency radio waves that propagate through the ionosphere before they reach ground-based receivers. The frequency of these signals is of the order of GHz and they therefore interact with small scale ionospheric irregularities (i.e. sharp gradients in ion and electron densities). This causes the signals to exhibit rapid amplitude and phase variations known as \emph{scintillations}, causing uncertainty in position and loss of lock in severe cases \cite{Kinter2007}. Interactions between the sun and the Earth’s ionosphere are extremely non-linear, which makes the prediction of space weather effects challenging. Because of the complex nature of the problem, a complete theory of ionospheric irregularities and signal scintillation does not yet exist, which limits the prediction capabilities of physics-based models~\cite{Priyadarshi2015}.

In this study, we focus on predicting phase scintillations at high latitudes, using data from Global Positioning System (GPS) receivers in the Canadian High Arctic Ionospheric Network (CHAIN) obtained between 2015-2017. At high latitudes, the dominant source of ionospheric irregularities and therefore scintillations is solar-driven storms and substorms (geomagnetically active periods). One very visible manifestation of these high energy inputs from the sun is the aurora, which has also been shown to correlate with these scintillations~\cite{aarons2000}~\cite{vandermeeren2015}. 

Phase scintillations are an uncommon yet severe phenomenon. Over the course of 2015-2016, only $0.0091\%$ of the minute basis samples from CHAIN exhibited scintillations over the threshold of $0.1$, which is the value above which the GNSS signal reliability decreases~\cite{ryan2018}. Therefore, the event task is sparse, rendering standard techniques incapable of predicting accurately in time and magnitude. In addition, due to errors in hardware, $22\%$ of data in the years 2015-2017 are missing, further complicating the prediction task.

We propose a method for predicting scintillations from time-series of physical measurements, incorporating two key novelties: (a) we account for the sparsity of scintillations with a custom loss function; (b) we handle missing data values with binary masks that inform our model which values are missing. We outperform the current state of the art by making predictions 1 hour in advance with a total skill score (TSS, discussed below) of $0.64$. To the best of our knowledge we are the first to treat the prediction problem as a regression problem, rather than classifying the existence of phase scintillations, as per the current state of the art.

\subsection{Related Work}
There is no physics based model capable of performing accurate predictions of timing and magnitude of phase scintillations with relevant spatial and temporal scales. Previous data-driven approaches include \cite{delima2015} where the authors predict scintillations in equatorial latitudes; we note that their results are not directly comparable to our task, as the physics guiding ionospheric scintillations on high latitudes differ from the equatorial. In \cite{jiao2013} the authors characterize the climatology of scintillations using statistical analysis but do not predict them in the future. The only known predictive model \cite{ryan2018} only classifies the occurrence of scintillations 1 hour in advance.
Our work is the first to treat phase scintillation prediction as a regression problem, and the first to account for event sparsity and missing data.

\section{Methodology}
\noindent\textbf{Dataset:} We study the high latitude Canadian sector ($50 \degree-70 \degree$ geographic latitude) between mid-October 2015 and end-February 2016. We use solar activity parameters such as solar wind speed (Vsw), interplanetary magnetic field components such as (IMF Bz, By), F10.7, Sym-H, etc. and geomagnetic activity indices such as Kp, AE to characterize the global influence of solar activity on Earth’s magnetosphere and ionosphere~\cite{spacephysics}. For local ionospheric state information pertaining to the high latitude Canadian sector, we are using the CARISMA (Canadian Array for Realtime Investigations of Magnetic Activity) magnetometer dataset, the Canadian High Arctic Ionospheric Network for ionospheric total electron content (TEC) and scintillation index measurements. In summary, we have $39$ features in a minute-cadence dataset. We use the first $75000$ points (until mid-January) as our training data, while we reserve the latter $75000$ points as our test set. 

\noindent\textbf{Masking:} Due to measurement errors, faulty equipment and other natural phenomena, $22\%$ of the data are corrupted and logged as \texttt{NaN}. For each feature, we substitute any NaN values with the mean value of that feature across all non-NaN observations. We indicate such substitutions with a binary (1/0) mask, which is provided as an additional feature. Including a mask for each feature therefore doubles the total number of features. This approach resembles that of \cite{Che2018}, except that we do not decay the substituted value towards the mean as the vast majority of our observations exhibit small variation from the mean.

\noindent\textbf{Sparsity and Loss Function:} As discussed above, phase scintillations are a rare phenomenon. Given the extreme sparsity of our positive phase scintillation values, standard regression loss functions would find their minimum by predicting values around the mean, and failing to predict high phase scintillation events.
We introduce a custom loss function defined in eq.\eqref{eq1}. The loss function has two components: (i) the Mean Absolute error (MAE) between the predicted output and ground truth; (ii) a dynamic range penalty that incentivizes the range of the output sequence to match the range of the true sequence.
With both components, the model is encouraged to match both the mean and variance of the true sequence. In the following expression for the loss $L$, the sum, max and min are computed over a batch $X, Y$ of predictions and ground truth respectively:

\begin{equation}
    L = \frac{\sum_i{\|y_i-x_i\|}}{N} + \lambda \left(\text{DynRange}(Y) - \text{DynRange}(X)\right)
\label{eq1}
\end{equation}
\begin{equation}
    \text{DynRange}(X) = \max(X)-\min(X)
\end{equation}

In our experiments we found that setting $\lambda = 0.1$ provided the best performance.

\begin{figure}
    \centering
    \includegraphics[width=.7\textwidth]{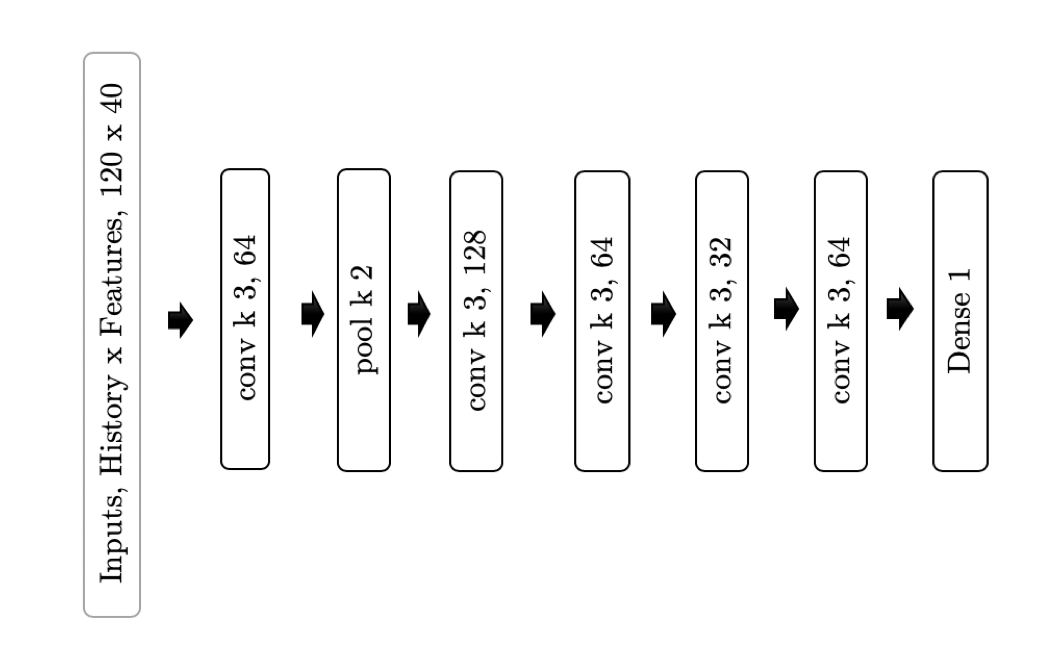}
    \caption{Proposed convolutional architecture. Input timeseries is passed through the convolutional layers and a single output neuron produces the prediction. {\tt{k}}: is the size of the convolutional kernel, while the following number corresponds to the filters}
    \label{arch}
\end{figure}

\noindent\textbf{Architecture:}
Figure \ref{arch} shows the proposed architecture. In addition to the time-series data for time $t$ we also include observations from $t-120$ min to $t$ min to inform the model of the historical changes in the values. Thus, the dimensionality of our input is $History \times Features$; in our case $120\times40$. We condition our model to predict 1 hour in advance. We note that filter sizes  decrease and then slightly increase in order to create a small bottleneck in feature space such that noisy information is excluded from the final layers. The aforementioned mask is appended to our input as an extra channel.

{}

\section{Results}
\begin{figure}
    \centering
    \includegraphics[width=.5\linewidth]{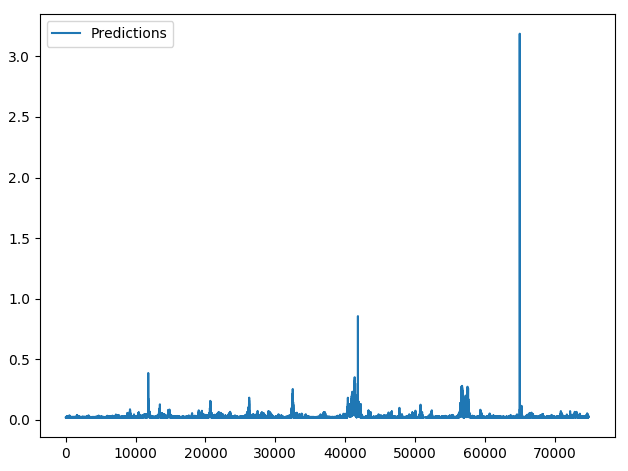}
    \caption{Predictions over 75000 points of test set; Note that predicted sequence exhibits same peak behaviour as ground truth.}
    \label{fig:my_label}
\end{figure}

Figure \ref{results} shows our results over the whole testing dataset of 75000 minute samples, as well as a zoomed version over 5000 minute samples. The test time-points do not overlap with the training time-points. It is evident that the prediction follows closely the ground truth but fails to accurately predict the magnitude of high intensity phase scintillations.
Fig \ref{results}(b) shows that despite the mismatch in absolute magnitude, our model successfully predicts a phase scintillation larger than the mean.
We also note a small delay in the order of 2-3 minutes in our predictions. Fig \ref{results}(a) shows that the predicted sequence has the same peak behaviour as the ground truth.
Fig.\ref{fig:my_label} shows the predictions over the whole test set to make the predicted time series more clear without the scaling of Fig. \ref{results}.

For ease of comparison against existing techniques such as \cite{ryan2018}, which approached the prediction task as one of classification, we quantitatively assess the performance of our model by first casting it as a binary classification task and then applying two classification metrics. To cast as binary classification, we say that a scintillation occurs if a threshold of 0.1 in the scintillation index is exceeded (this choice of threshold follows that chosen in \cite{jiao2013}). The metrics we use to assess this are the total skill score (TSS) as defined in \cite{ryan2018} and eq.~\ref{eq2}, and the Heidke skill score as discussed in \cite{heidtke}. The Heidke skill score takes values in $(-\inf,1]$, and the TSS takes values in $[-1, 1]$. In both cases, a score of 0 means no predictive capability.
Our model scores $0.34$ on the Heidke skill score, showing some predictive ability.
The model of \cite{ryan2018} performed classification with a TSS of $0.49$, while our model achieves a TSS of $0.64$, outperforming the current state of the art.
\begin{equation}
    TSS= \frac{TP}{(TP + FN)} - \frac{FP}{(FP + TN)}
    \label{eq2}
\end{equation}
\begin{figure}
\centering
\begin{minipage}{.5\textwidth}
  \centering
  \includegraphics[width=1\linewidth]{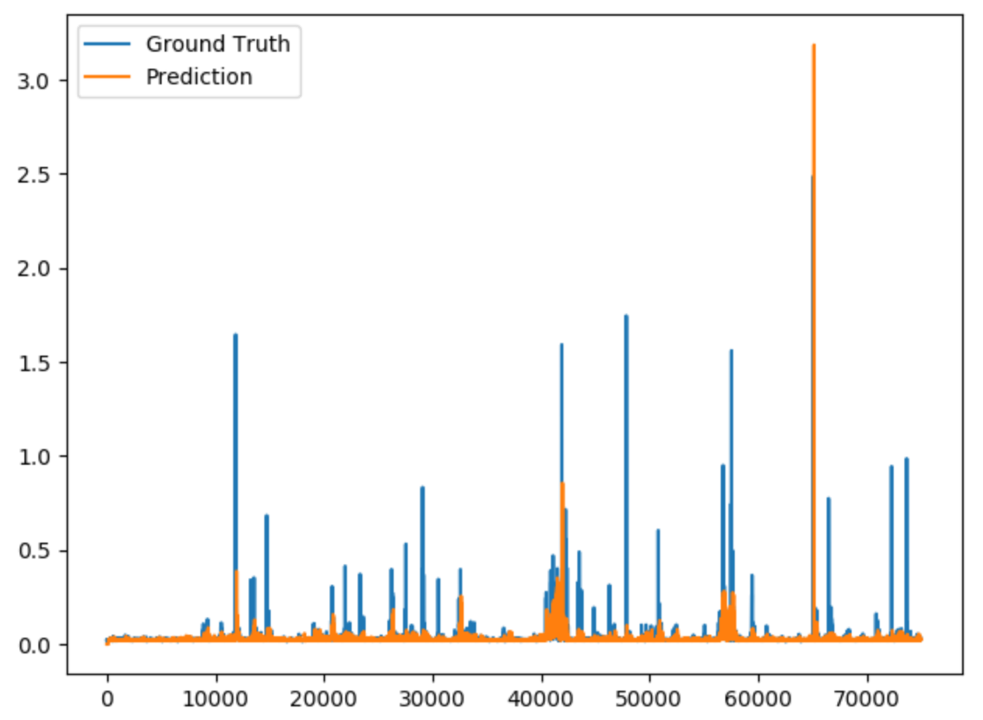}
  \label{fig:test1}
\end{minipage}%
\begin{minipage}{.5\textwidth}
  \centering
  \includegraphics[width=1\linewidth]{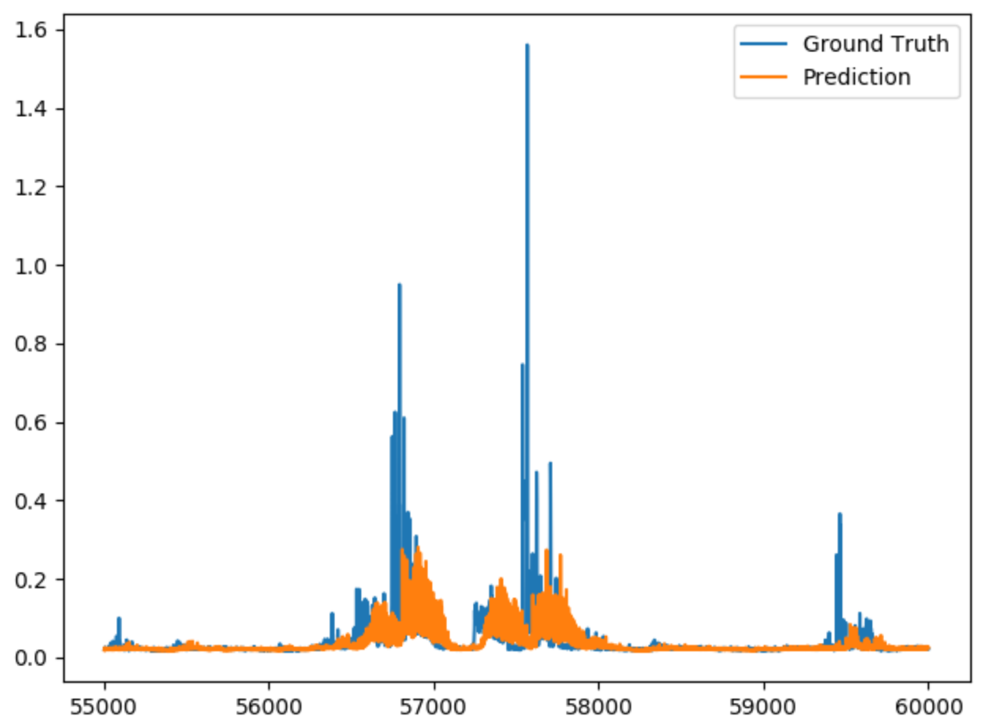}
  \label{fig:test2}
\end{minipage}
\label{results}
 \caption{(a) Results over 75000 minute interval (b) Results zoomed in a 5000 min interval}
\label{results}
\end{figure}

{}
\section{Conclusions}
GNSS products like GPS are vital to modern day operations; from navigation to high-frequency trading in the stock markets --- all of which can be disrupted by scintillations in the signals.
In this paper we proposed a novel methodology for predicting GNSS phase scintillations 1 hour in advance. We introduced a custom loss function and a method of dealing with missing and incomplete data. We further improved the current state of the art by $0.15$ in skill score and demonstrated our method's predictive capability.
We believe that additional data sources driven by the same physical phenomena, like auroral images, can further improve the skill score.

\bibliographystyle{splncs03}
\bibliography{bibli}
\end{document}